%% file: main.tex
\documentclass[11pt,letterpaper]{article}
\usepackage[T1]{fontenc}
\usepackage{lmodern}
\usepackage[hmargin=0.9in,top=0.83in,bottom=0.9in]{geometry}
\usepackage{amsmath,amssymb,graphicx,booktabs,adjustbox,xspace,microtype}
\usepackage{placeins,xcolor,needspace,float}
\usepackage[font=small,labelfont=bf,labelsep=period]{caption}
\usepackage[numbers,sort&compress]{natbib}
\usepackage[hidelinks]{hyperref}
\usepackage[capitalize,nameinlink,noabbrev]{cleveref}

\newcommand{\ours}{GGDriver\xspace}
\newcommand{\cmark}{\ensuremath{\checkmark}}
\hypersetup{pdftitle={GGDriver: Glance at What Is Shared, Gaze at What Differs for Perception-Free Planning},pdfauthor={Jiawei Sun, Maosen Li, Jiamin Shi, Xiaoxu Zheng, Jiun Tian Hoe, Yechao Bai, Michael Bi Mi}}
\title{\bfseries\Large GGDriver: Glance at What Is Shared, Gaze at What Differs\\
for Perception-Free Planning}
\author{\normalsize Jiawei Sun \quad Maosen Li \quad Jiamin Shi \quad Xiaoxu Zheng\\[3pt]
\normalsize Jiun Tian Hoe \quad Yechao Bai \quad Michael Bi Mi}
\date{}
\begin{document}
\maketitle
\input{sections/paper}
\FloatBarrier
{\small
\bibliographystyle{unsrtnat}
\bibliography{references}
}
\end{document}

%% file: sections/paper.tex
\begin{abstract}
A planner deciding whether to turn left or continue straight needs to inspect different parts of the scene for each maneuver.
Yet many end-to-end driving planners evaluate every candidate trajectory against the same compressed scene representation.
We let each candidate select the visual evidence used to refine and score it.
A global glance first produces trajectory hypotheses from shared scene features.
We then project a corridor around each trajectory into the camera views and use the resulting mask to attend to patch features at the backbone's native resolution.
After refinement, the scorer projects each final proposal again and ranks it using evidence along its own path.
This geometric selection requires only the predicted trajectory and camera calibration.
On three planners with different scene representations (learned registers, a perception-supervised bird's-eye-view grid, and a coarse image-token grid), adding this gaze improves planning scores.
In ablations, path-aligned patches outperform both a matched number of randomly selected patches and attention over all patches.
Using the gaze for ranking yields larger gains than using it for refinement alone.
With a ViT-S backbone and no perception labels, GGDriver achieves a rater feedback score of 8.110 on the WOD-E2E test set, 94.2 PDMS on NAVSIM-v1, 48.6 EPDMS on NAVSIM-v2, and a closed-loop Driving Score of 72.18 on Bench2Drive.
\end{abstract}

\section{Introduction}
\label{sec:intro}

\begin{figure}[htbp]
  \centering
  \includegraphics[width=0.78\linewidth]{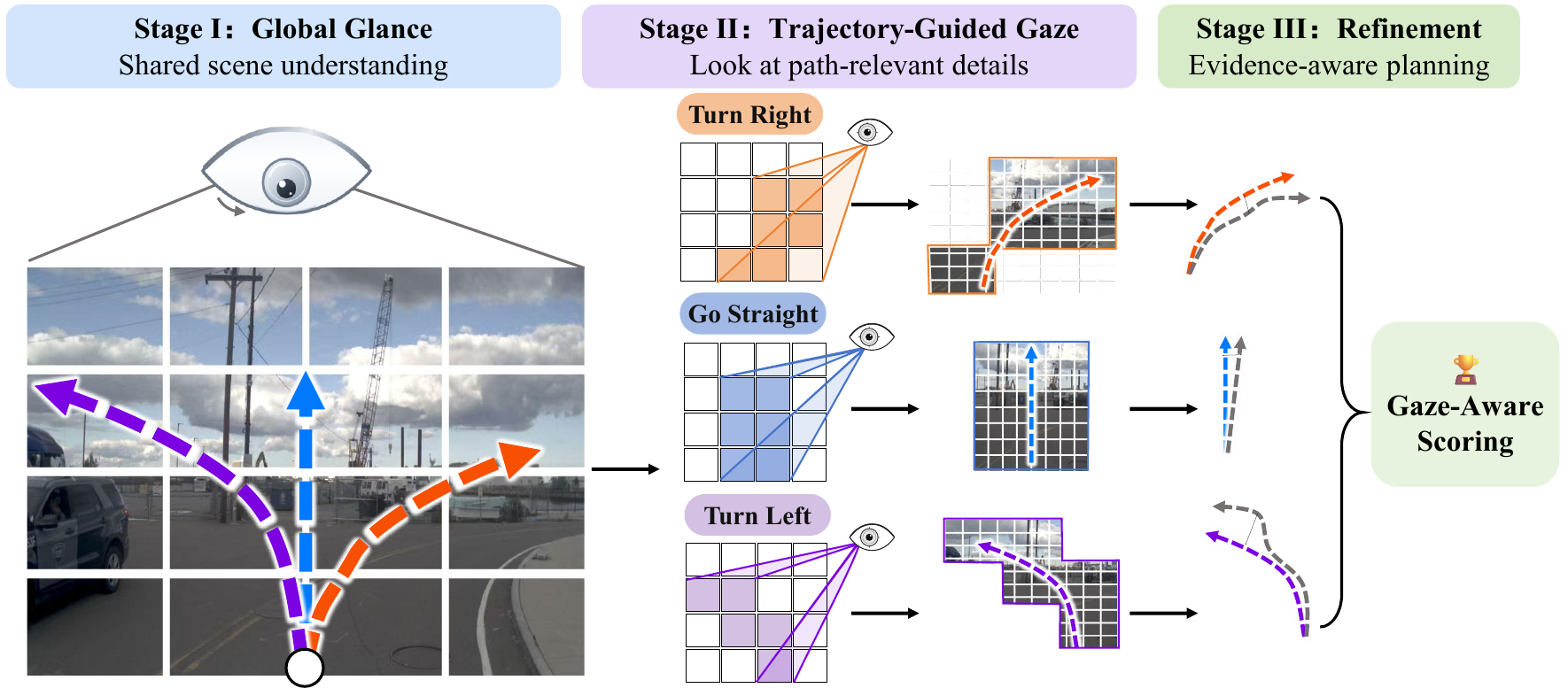}

  \caption{\textbf{Glance at what is shared, gaze at what differs.}
  A glance at a coarse view of the scene, shared by all planning queries, yields the trajectory hypotheses (left), here a right turn, a straight continuation and a left turn.
  Each hypothesis then gazes at the full-resolution patches of the image blocks its own path passes through (middle).
  This evidence refines each hypothesis and conditions the scorer that ranks them (right): the hypothesis decides which evidence is acquired.}
  \label{fig:teaser}
\end{figure}

Human drivers do not inspect a scene uniformly: they fixate on the road they are about to steer into~\cite{land1994where}, and a peripheral glance is enough for the rest.
What they look at depends on what they intend to do.
End-to-end planners~\cite{hu2023uniad,jiang2023vad,weng2024paradrive,sun2024sparsedrive,liao2025diffusiondrive} typically work the other way round: the scene is compressed into a fixed representation before any plan exists, and planning only consumes it.
Most of them build a bird's-eye-view (BEV) or sparse scene representation and supervise it with 3D boxes, HD-map elements and motion forecasts.
These labels are expensive, they tie the model to the sensor setup and taxonomy of one dataset, and they shape the representation around what an annotator chose to label rather than what the planner needs.

\emph{Perception-free} planners drop these labels and learn from trajectories, and optionally from trajectory scores, only~\cite{li2024hydramdp,kirby2026drivor}.
The recent DrivoR~\cite{kirby2026drivor} shows how far this can go.
It appends a few learnable register tokens~\cite{darcet2024registers} to a pretrained ViT~\cite{dosovitskiy2021vit,oquab2024dinov2}, and a light trajectory decoder and scorer read only these registers.
With 16 registers per camera it matches or exceeds perception-heavy planners on NAVSIM~\cite{dauner2024navsim}.

Whatever the representation, BEV grid, sparse queries or registers, it is built once, before any plan exists, and \emph{every} candidate trajectory reads the \emph{same} summary.
Two properties of driving make this a poor trade-off.
First, \textbf{different plans need different evidence}.
A hypothesis that turns left must check the crosswalk at the turn exit, while one that keeps straight must check the far lane; a shared summary has to encode both, plus everything else, in the same few tokens.
Second, \textbf{decisive cues are small}.
A pedestrian 40\,m ahead or the gap in front of a stopped vehicle covers a few patches, which are easily lost when a camera is pooled into a few tokens.
On long-tail benchmarks such as WOD-E2E~\cite{xu2025wode2e}, such small, plan-specific cues often separate a good plan from a bad one.
Progress has largely come from better global representations; we argue that this misses an orthogonal axis.

Planning should not merely consume visual evidence: its hypotheses should help determine which evidence to acquire.
Like a human driver, a planner can first form a hypothesis of where to go and then acquire the evidence this hypothesis depends on, which lies largely along its own path.
In driving, the hypothesis already says where this evidence is: a trajectory in the ego frame and the camera calibration determine which image regions the vehicle is about to drive through, so the hypothesis itself is the attention mask.
We call this \emph{hypothesis-specific evidence acquisition} and summarize it as \emph{glance at what is shared, gaze at what differs} (\cref{fig:teaser}):
\begin{itemize}
  \item \textbf{Glance.} All hypotheses read the planner's global scene representation, whatever it is, and predict a trajectory. The glance carries what the hypotheses have in common, such as the road layout and the overall traffic situation.
  \item \textbf{Gaze.} The corridor of each predicted trajectory is projected into the cameras, and the hypothesis attends to the full-resolution patch features of the image regions it will traverse. The gaze reads what differs between the hypotheses: the evidence along each one's own path.
\end{itemize}
In the trajectory decoder, the acquired evidence \emph{refines} each hypothesis. Crucially, evidence acquisition does not end with refinement: the scorer gazes again along each final proposal, so that every candidate is \emph{ranked} on the evidence along its own path, rather than on a summary shared by all or on features accumulated during refinement.
No observation policy is learned: there is no selection network and no label for it.

Because the gaze only needs a trajectory and the camera calibration, it is independent of the glance, and we test it as a mechanism rather than as one architecture.
Added to planners with three different global representations, learned registers (DrivoR~\cite{kirby2026drivor}), a perception-supervised BEV (DiffusionDrive~\cite{liao2025diffusiondrive}) and a coarse token grid, it improves all of them.
Controlled ablations show that the gain comes from \emph{where} each hypothesis looks, not from reading more: random regions of the same size hurt, and reading every region is worse than reading the path.
They also show that evidence-conditioned ranking, more than refinement, carries the gain.
\ours, the coarse token grid with the gaze, achieves the highest WOD-E2E test RFS among the methods in \cref{tab:wod} and is competitive with the state of the art on NAVSIM-v1 and NAVSIM-v2.

\Needspace{4\baselineskip}
Our contributions are as follows:
\begin{enumerate}
  \item We identify hypothesis-specific evidence acquisition as an axis of perception-free planning that is orthogonal to the global scene representation, and realize it with a gaze that is selected by projecting each trajectory hypothesis into the cameras, with no selection network, no extra labels and no perception supervision.
  \item We use the acquired evidence to refine hypotheses and to rank them. This \emph{evidence-conditioned ranking}, which scores each candidate on the evidence along its own path, contributes the larger share of the gain.
  \item We show that the mechanism is general: it improves planners built on registers, on a BEV and on a token grid, and controlled ablations attribute the gain to where a hypothesis looks rather than how much it reads. \ours achieves the highest WOD-E2E test RFS among the methods in \cref{tab:wod}, is competitive with the state of the art on NAVSIM-v1 and NAVSIM-v2, and reaches a closed-loop Driving Score of 72.18 on Bench2Drive.
\end{enumerate}

\section{Related Work}
\label{sec:related}

\noindent\textbf{Perception-supervised end-to-end planning.}
UniAD~\cite{hu2023uniad} chains detection, tracking, mapping, motion forecasting and occupancy prediction into one planning-oriented network, and VAD~\cite{jiang2023vad}, PARA-Drive~\cite{weng2024paradrive} and SparseDrive~\cite{sun2024sparsedrive} make this pipeline vectorized, parallel and sparse.
TransFuser~\cite{chitta2023transfuser} and its camera-only variant LTF fuse sensors in a BEV grid with auxiliary semantic and detection heads.
DiffusionDrive~\cite{liao2025diffusiondrive} denoises a predefined set of anchor trajectories that query BEV features at the trajectory locations.
iPad~\cite{guo2025ipad} refines sparse, proposal-centric BEV queries: the corners of each proposal are lifted to 3D pillars, deformable attention samples the multi-view features at learned offsets, and an MLP scores the pooled proposal features.
These methods show that \emph{trajectory-anchored} feature reading helps planning, but all of them rely on perception supervision, either through the BEV they sample from or through auxiliary heads (iPad, for example, predicts on-road probabilities and the future of likely-to-collide agents).
\ours needs neither a BEV representation nor learned sampling offsets: its hypotheses are regressed from learned queries without a trajectory vocabulary, the corridor of each one selects native-resolution patches through a fixed geometric mask, and the scorer reads the patches along each final proposal again instead of ranking the features accumulated during refinement.

\noindent\textbf{Perception-free and scoring-based planning.}
Since ego-status shortcuts were found in open-loop evaluation~\cite{li2024egostatus}, benchmarks score plans by driving quality~\cite{dauner2024navsim,cao2025pseudosim,xu2025wode2e,jia2024bench2drive}, and planners learn to score their own candidates.
Hydra-MDP~\cite{li2024hydramdp,li2025hydramdppp} distills rule-based metric scores into a scorer over a trajectory vocabulary, DriveSuprim~\cite{yao2025drivesuprim} refines the selection from coarse to fine, and LAW~\cite{li2025law} replaces perception labels with a latent world-model objective.
DrivoR~\cite{kirby2026drivor} is closest to our work: it compresses each camera into 16 ViT registers, and decodes and scores 64 trajectory proposals against these shared tokens.
We keep its perception-free training, its losses and its decoders, and add what a shared summary cannot provide: a per-hypothesis, full-resolution gaze along the hypothesis' own path, in the trajectory decoder and in the scorer.

\noindent\textbf{Adaptive visual reading.}
Learned queries that compress large inputs, as in Perceiver~\cite{jaegle2021perceiver}, the Q-Former of BLIP-2~\cite{li2023blip2} and ViT registers~\cite{darcet2024registers}, are shared by all downstream consumers, and token sparsification~\cite{rao2021dynamicvit} drops patches once for the whole model.
Deformable attention~\cite{zhu2021deformable} lets each query sample a few learned offsets, but it gives no explicit signal about \emph{what} a query should look at.
Active perception~\cite{bajcsy1988active} and the recurrent attention model~\cite{mnih2014ram} decide where to look next with a designed or learned policy; \ours has no such policy, because the hypothesis under consideration decides what is read.
Every hypothesis of \ours has its own gaze in every decoder layer and in the scorer, and the image regions it reads are given by its own camera projection rather than learned.

\section{Method}
\label{sec:method}

\begin{figure}[htbp]
  \centering
  \includegraphics[width=\linewidth]{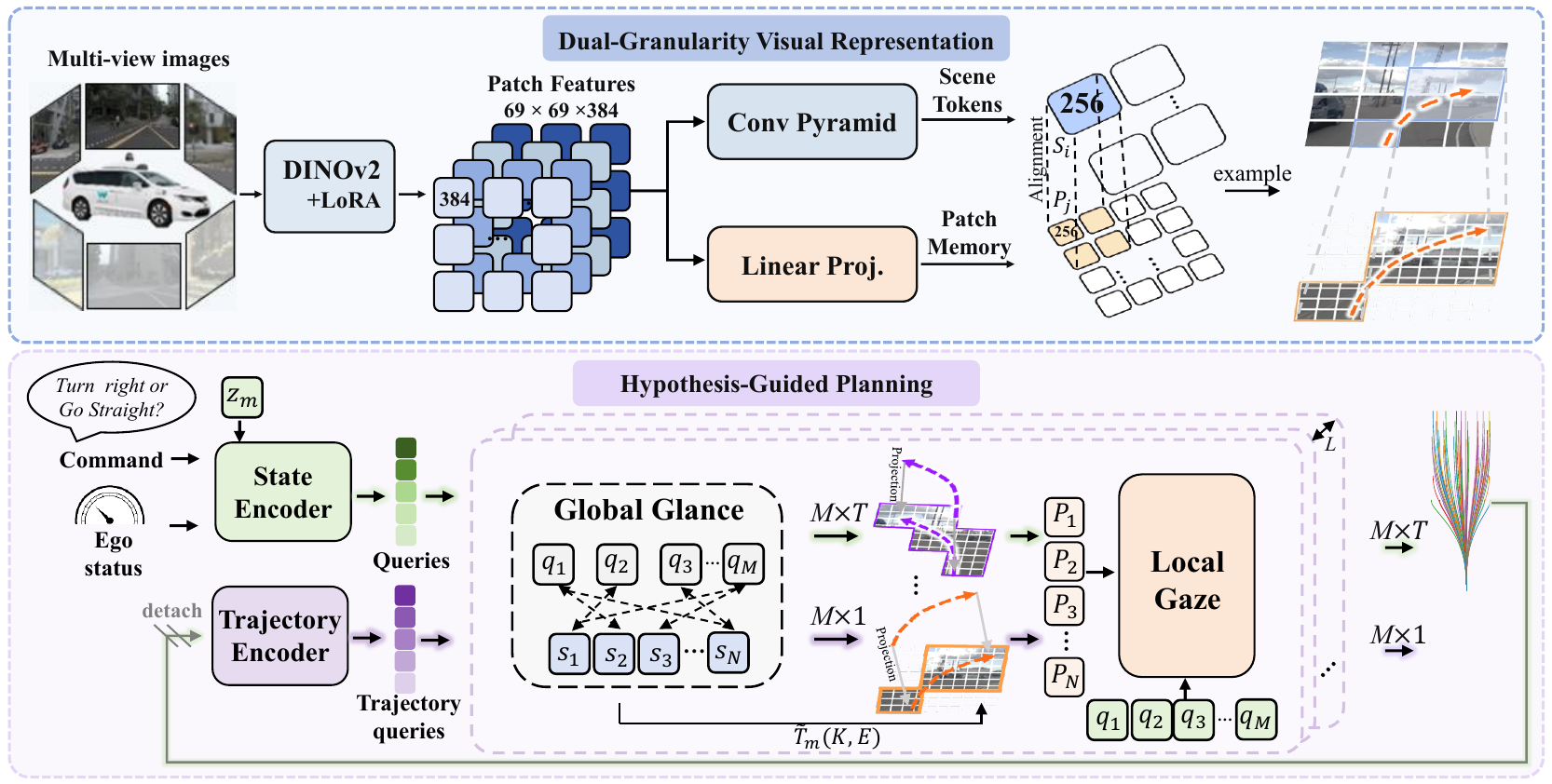}
  \caption{\textbf{Overview of \ours.} Every decoder layer performs a global glance over shared scene tokens, predicts a trajectory, projects its corridor into the cameras, and gazes at the full-resolution patches of the covered blocks. The scorer ranks each proposal after the same gaze along that proposal's path. The projection needs only the trajectory and the camera calibration. Sizes are per camera, height $\times$ width $\times$ channels.}
  \label{fig:arch}
\end{figure}

\cref{fig:arch} gives an overview.
\ours takes $V$ calibrated camera images and the ego status (velocity, acceleration and driving command).
It outputs $M$ trajectory proposals $\{\mathbf{T}_m\}_{m=1}^M$ with $\mathbf{T}_m\in\mathbb{R}^{H\times 3}$ (position and heading over $H$ future steps), together with a score for each proposal.

\subsection{Scene tokens and patch memory}
\label{sec:method:encoder}

Each image $I_v$ passes through a DINOv2 ViT-S/14~\cite{oquab2024dinov2} fine-tuned with LoRA~\cite{hu2022lora}.
We discard the class and register tokens and keep the patch features $F_v\in\mathbb{R}^{H_p\times W_p\times C}$.
The trajectory decoder and the scorer read them in two forms: the patches themselves for the gaze, and a compact set of scene tokens for the glance.

\noindent\textbf{Patch memory.}
A linear layer projects $F_v$ to dimension $D$, and we add learned patch-position and camera embeddings.
We stack the result into $P\in\mathbb{R}^{VH_pW_p\times D}$.
The patch memory has no spatial downsampling, so a gaze reads the backbone's native resolution.
To address it by image region, we partition the patches of every image into a $G\times G$ grid of non-overlapping blocks, each covering about $h\times w\approx(H_p/G)\times(W_p/G)$ patches.
We index the $N=VG^2$ blocks of all cameras by $j$, and write $P_j$ for the keys of block $j$ and $b(i)$ for the block that contains key $i$.

\noindent\textbf{Glance tokens.}
The glance needs a summary of all cameras that is compact enough for every hypothesis to read in full.
By default, a light convolutional pyramid $\phi$ (stride-2 convolutions followed by adaptive pooling) reduces $F_v$ to one token per block:
\begin{equation}
  \mathbf{s}_{v,b} = \phi(F_v)_b + \mathbf{e}^{\text{scene}}_{v,b}, \qquad b\in\{1,\dots,G^2\},
\end{equation}
where $\mathbf{e}^{\text{scene}}_{v,b}$ is a learned embedding per camera and block.
We stack the tokens of all cameras into $S\in\mathbb{R}^{N\times D}$ and write $\mathbf{s}_j$ for the token of block $j$.
With $G=4$ and $V=4$, $N=64$, the same number of scene tokens as DrivoR's 16 registers per camera, and each token is $hw$ times coarser than the patches of its block (about 289$\times$ with our WOD-E2E setting).
This token grid is one implementation choice: the gaze addresses the patch memory, not the scene tokens, and it works with DrivoR's registers as the glance as well (\cref{sec:exp:ablation}).

\subsection{Glance-and-gaze decoder}
\label{sec:method:decoder}

Following~\cite{kirby2026drivor}, the initial queries are $\mathbf{q}^0_m=W_e\,\mathbf{e}_{\text{ego}}+\mathbf{z}_m$, where $\mathbf{e}_{\text{ego}}$ is the ego status and $\mathbf{z}_m$ are $M$ learned mode embeddings.
An MLP head gives the initial trajectories $\mathbf{T}^0_m$; there are no anchor trajectories and no trajectory vocabulary.
Each of the $L$ decoder layers then performs the three steps below.
We use the superscript $\ell$ for the output of layer $\ell$.

\noindent\textbf{Glance.}
The queries attend to each other and to all scene tokens, and an MLP head $h^{\ell}_{g}$ predicts a trajectory from the result:
\begin{align}
  \tilde{\mathbf{q}}^{\ell}_m &= \operatorname{FFN}\big(\operatorname{CA}(\operatorname{SA}(\mathbf{q}^{\ell-1}),\, S)\big)_m, \\
  \tilde{\mathbf{T}}^{\ell}_m &= h^{\ell}_{g}\big(\tilde{\mathbf{q}}^{\ell}_m\big),
\end{align}
where SA, CA and FFN are pre-norm self-attention, cross-attention and feed-forward blocks with residual connections, exactly the decoder layer of DrivoR.
The glance forms a coarse driving intention from the whole surround view, and $\tilde{\mathbf{T}}^{\ell}_m$ states where this intention leads.

\noindent\textbf{Projection.}
The corridor of $\tilde{\mathbf{T}}^{\ell}_m$ is projected into the cameras (\cref{sec:method:corridor}).
This gives the set $\Omega^{\ell}_m\subseteq\{1,\dots,N\}$ of image blocks the hypothesis is about to drive through.
The projection is a fixed geometric function of the trajectory and the calibration, with no parameters and no gradient.

\noindent\textbf{Gaze and trajectory update.}
Each query attends only to its own blocks, and the trajectory is predicted again:
\begin{align}
  \mathbf{q}^{\ell}_m &= \operatorname{FFN}\Big(\operatorname{CA}\big(\tilde{\mathbf{q}}^{\ell}_m,\, \{P_j\}_{j\in\Omega^{\ell}_m}\big)\Big), \\
  \mathbf{T}^{\ell}_m &= h^{\ell}\big(\mathbf{q}^{\ell}_m\big).
\end{align}
We implement the gaze as one masked attention over the full patch memory, with a boolean mask $\mathbf{1}[b(i)\in\Omega^{\ell}_m]$ for query $m$ and key $i$; a hypothesis whose corridor leaves every camera skips the attention term.
The updated query enters the next layer, where the next glance predicts a new trajectory and therefore a new gaze.
This closes a loop in which the hypothesis decides which evidence is read, and the evidence corrects the hypothesis.

\noindent\textbf{Why the trajectory, and not a learned selector.}
One could instead train a network to score blocks and keep the top-$k$.
Such a selector needs a target, and the only label-free target is again the projected trajectory, so it can at best imitate the projection; it also cannot be trained through the planning loss, because top-$k$ selection is not differentiable.
Using the trajectory directly removes the selector, its loss, its hyperparameters and the gap between its training target and its test-time behavior.
The glance trajectory $\tilde{\mathbf{T}}^{\ell}_m$ is supervised like every other trajectory output (\cref{sec:method:loss}), so the gaze improves together with the planner.

\subsection{Corridor projection}
\label{sec:method:corridor}

Given a trajectory $\mathbf{T}=\{(x_t,y_t,\theta_t)\}_{t=1}^{H}$ in the ego frame, we start its path at the ego origin, extend it along its final heading until it is at least $L_{\min}$ long, and resample it every $\Delta$ meters, so that the corridor does not skip blocks between distant waypoints at high speed.
The extension matters for a stationary or slow ego vehicle, whose 4--5\,s trajectory may cover a few meters while the relevant scene, a crossing pedestrian or the vehicle in front, is further ahead.
For every point $(x,y,\theta)$ of this path we form corridor points at lateral offsets $\delta\in\{-\tfrac{w_c}{2},0,\tfrac{w_c}{2}\}$ and heights $z\in\{0,1\}$\,m,
\begin{equation}
  \mathbf{c}_{\delta,z} = \big(x - \delta\sin\theta,\; y + \delta\cos\theta,\; z\big),
\end{equation}
with a corridor width $w_c$ slightly larger than the vehicle.
The ground points mark the road the hypothesis drives on, and the points at 1\,m height mark the bodies of vehicles and pedestrians that stand in its way.
We project each point into camera $v$ with the intrinsics $K_v$ and the ego-to-camera extrinsics $E_v$, written $\pi_v$, and a block $j$ belongs to $\Omega$ if at least one projection with depth $d>d_{\min}$ falls inside it:
\begin{equation}
  \Omega(\mathbf{T}) = \big\{\, j \;:\; \exists\, \mathbf{c}_{\delta,z}\ \text{with}\ d>d_{\min}\ \text{and}\ \pi_v(\mathbf{c}_{\delta,z})\in\text{block } j \,\big\}.
\end{equation}

\subsection{Evidence-conditioned ranking}
\label{sec:method:scorer}

DrivoR ranks the final proposals with a scoring decoder that embeds $\operatorname{sg}[\mathbf{T}^L_m]$ (stop-gradient), attends to the scene tokens over $L_s$ layers, and predicts a quality score per proposal.
All candidates are thus ranked against the same compressed summary: the scorer must judge whether a proposal is safe without reading the pixels along it.
We instead condition the score of each candidate on its own evidence.
The proposals are fixed inside the scorer, so their corridors $\Omega(\mathbf{T}^L_m)$ are projected once, and each scorer layer performs its global cross-attention followed by a masked cross-attention to the patches of $\Omega(\mathbf{T}^L_m)$.
The scorer remains a learned ranking function, trained as before (\cref{sec:method:loss}): it checks no explicit rule, but ranks the candidates given what each of them would drive through.
At inference time we output the highest-scoring proposal.
This is the more effective of the two uses of the gaze (\cref{sec:exp:ablation}).

\subsection{Training objective}
\label{sec:method:loss}

We keep the losses of DrivoR and add one term for the glance trajectories.
For each decoder layer $\ell$, a winner-take-all loss~\cite{chai2019multipath} trains the proposals toward the expert trajectory $\mathbf{T}^{*}$, and the glance trajectories receive the same loss, because they decide where the layer gazes:
\begin{equation}
  \mathcal{L}_{\text{traj}} = \sum_{\ell=1}^{L}\Big( \min_m \big\|\mathbf{T}^{\ell}_m-\mathbf{T}^{*}\big\|_1 + \min_m \big\|\tilde{\mathbf{T}}^{\ell}_m-\mathbf{T}^{*}\big\|_1 \Big).
\end{equation}
The total loss is $\mathcal{L}=\mathcal{L}_{\text{traj}}+\lambda_{\text{score}}\mathcal{L}_{\text{score}}$, where the scoring loss is a binary cross-entropy against the benchmark metric of each proposal: the PDM sub-scores on NAVSIM~\cite{kirby2026drivor} and the normalized rater feedback score on WOD-E2E.
The network never receives boxes, maps or semantic labels as supervision; on NAVSIM, the metric targets of the scorer are computed by the benchmark's own evaluation code, as in prior scoring-based planners~\cite{li2024hydramdp,kirby2026drivor}.
Apart from this term, the gaze adds no loss and no label.

\section{Experiments}
\label{sec:exp}

\subsection{Setup}
\label{sec:exp:setup}

\noindent\textbf{Benchmarks.}
\emph{WOD-E2E}~\cite{xu2025wode2e} contains 4{,}021 long-tail segments recorded with eight surround cameras and evaluates 5\,s plans with the rater feedback score (RFS, range 4--10), which rewards trajectories inside trust regions around rater-preferred trajectories; we also report ADE.
\emph{NAVSIM-v1}~\cite{dauner2024navsim} evaluates 4\,s plans on \texttt{navtest} with the PDM score (PDMS), which combines no-at-fault collision (NC), drivable-area compliance (DAC), time-to-collision (TTC), comfort (Comf.) and ego progress (EP).
\emph{NAVSIM-v2}~\cite{cao2025pseudosim} uses pseudo-simulation with a second, reactive stage on the challenging \texttt{navhard} split and reports the extended score EPDMS.
\emph{Bench2Drive}~\cite{jia2024bench2drive} drives 220 short routes in CARLA in closed loop and reports the Driving Score (DS), the success rate (SR), efficiency, comfort and the success rate per driving ability.

\noindent\textbf{Implementation.}
Following DrivoR~\cite{kirby2026drivor}, we use a DINOv2 ViT-S/14 backbone fine-tuned with LoRA and DrivoR's training configuration; the only change is the glance and the gaze.
On every benchmark, the model takes $V=4$ cameras as input (front, front-left, front-right and rear); on WOD-E2E, these are four of the eight recorded cameras.
The block grid is $G=4$, the trajectory decoder and the scorer have $L=L_s=4$ layers and $M=64$ hypotheses, and all benchmarks use the same corridor ($w_c=3$\,m), which covers 8.4 of the 64 blocks per hypothesis on average.
Bench2Drive uses the same setup as WOD-E2E.

\subsection{Main results}
\label{sec:exp:main}

\ours is a pure imitation-learning planner, without a language model, a world model, a reward model or perception labels; \cref{tab:wod,tab:navsim_v1,tab:navsim_v2} mark the paradigm of every compared method.

\begin{table}[htbp]
  \centering\small
  \setlength{\tabcolsep}{4pt}
  \begin{adjustbox}{max width=\linewidth}
\begin{tabular}{l l c ccc cccc}
    \toprule
    & & & & \multicolumn{2}{c}{ADE$\downarrow$} & \multicolumn{4}{c}{RFS by scenario$\uparrow$} \\
    \cmidrule(lr){5-6}\cmidrule(lr){7-10}
    Method & Venue & Type & RFS$\uparrow$ & 5\,s & 3\,s & Constr. & Inters. & Ped. & Cyc. \\
    \midrule
    DrivePI~\cite{liu2025drivepi} & CVPR'26 & VLA & 7.516 & 2.907 & 1.253 & 8.249 & 7.830 & 7.929 & 7.703 \\
    AutoVLA~\cite{zhou2025autovla} & NeurIPS'25 & VLA+RL & 7.557 & 2.958 & 1.351 & 7.956 & 7.711 & 7.592 & 7.321 \\
    IRL-VLA~\cite{jiang2025irlvla} & CVPR-F'26 & VLA+WM+RL & 7.890 & 2.824 & 1.218 & 8.440 & 7.978 & 7.924 & 7.776 \\
    Poutine~\cite{rowe2025poutine} & arXiv'25 & VLA+RL & 7.986 & 2.742 & 1.206 & 8.360 & 8.136 & 7.953 & 7.833 \\
    DriveMA~\cite{zheng2026drivema} & CoRL'26 & VLA+RL & 8.075 & 2.662 & \textbf{1.163} & 8.595 & 8.108 & 8.055 & \textbf{7.893} \\
    \midrule
    DrivoR$^\dagger$~\cite{kirby2026drivor} & CVPR'26 & IL & 7.783 & 2.956 & 1.265 & 8.650 & 7.875 & 7.848 & 7.605 \\
    FROST-Drive~\cite{dong2026frostdrive} & WACVW'26 & IL & 7.856 & 3.565 & 2.537 & 8.251 & 7.966 & 7.929 & 7.719 \\
    NTR~\cite{li2026ntr} & CoRL'26 & IL & 7.998 & 2.755 & 1.190 & \textbf{8.949} & 8.106 & 7.978 & 7.813 \\
    RAP (ensemble)~\cite{feng2026rap} & ICLR'26 & IL & 8.043 & \textbf{2.646} & 1.174 & 8.694 & 8.180 & 8.034 & 7.760 \\
    \midrule
    \textbf{\ours (ours)} & -- & IL & \textbf{8.110} & 2.682 & 1.213 & 8.768 & \textbf{8.224} & \textbf{8.100} & 7.854 \\
    \bottomrule
  \end{tabular}
  \end{adjustbox}
  \caption{\textbf{WOD-E2E test set.} VLA: vision-language-action model, WM: world model, RL: reinforcement learning, IL: imitation learning only. $^\dagger$Our reproduction in our setting. Bold: best.}
  \label{tab:wod}
\end{table}

\noindent\textbf{WOD-E2E.}
\ours reaches 8.110 RFS on the test set (\cref{tab:wod}), the highest RFS among the methods compared here.
With a ViT-S backbone and imitation learning only, it surpasses each VLA model in the comparison, including those post-trained with reinforcement learning or trained against a learned reward model.
Among imitation-learning planners it is ahead, as a single model, of the two-checkpoint ensemble of RAP, and it is +0.33 above our DrivoR reproduction.
It also has the best intersection and pedestrian scores, the clusters in which the decisive road user usually stands on or next to the planned path.
Its 5\,s ADE is within 0.04\,m of the best; RFS is the official ranking metric.

\begin{table}[htbp]
  \centering\small
  \setlength{\tabcolsep}{4pt}
  \begin{adjustbox}{max width=\linewidth}
\begin{tabular}{l l c ccccc c}
    \toprule
    Method & Venue & Type & NC$\uparrow$ & DAC$\uparrow$ & TTC$\uparrow$ & Comf.$\uparrow$ & EP$\uparrow$ & PDMS$\uparrow$ \\
    \midrule
    Human & -- & -- & 100 & 100 & 100 & 99.9 & 87.5 & 94.8 \\
    PDM-Closed~\cite{dauner2023pdm} & CoRL'23 & privileged & 94.6 & 99.8 & 86.9 & 99.9 & 89.9 & 89.1 \\
    \midrule
    UniVLA~\cite{wang2025univla} & ICLR'26 & VLA+WM & 96.9 & 91.1 & 91.7 & 96.7 & 76.8 & 81.7 \\
    DrivingGPT~\cite{chen2025drivinggpt} & ICCV'25 & WM & 98.9 & 90.7 & 94.9 & 95.6 & 79.7 & 82.4 \\
    AutoVLA~\cite{zhou2025autovla} & NeurIPS'25 & VLA+RL & 98.4 & 95.6 & \textbf{98.0} & 99.9 & 81.9 & 89.1 \\
    DriveVLA-W0~\cite{li2026drivevlaw0} & ICLR'26 & VLA+WM & 98.7 & \textbf{99.1} & 95.3 & 99.3 & 83.3 & 90.2 \\
    ReCogDrive~\cite{li2026recogdrive} & ICLR'26 & VLA+RL & 97.9 & 97.3 & 94.9 & \textbf{100} & 87.3 & 90.8 \\
    \midrule
    UniAD~\cite{hu2023uniad} & CVPR'23 & IL & 97.8 & 91.9 & 92.9 & \textbf{100} & 78.8 & 83.4 \\
    LTF~\cite{chitta2023transfuser} & TPAMI'23 & IL & 97.4 & 92.8 & 92.4 & \textbf{100} & 79.0 & 83.8 \\
    DiffusionDrive~\cite{liao2025diffusiondrive} & CVPR'25 & IL & 98.2 & 96.2 & 94.7 & \textbf{100} & 82.2 & 88.1 \\
    Hydra-MDP++~\cite{li2025hydramdppp} & arXiv'25 & IL & 98.6 & 98.6 & 95.1 & \textbf{100} & 85.7 & 91.0 \\
    DriveSuprim~\cite{yao2025drivesuprim} & AAAI'26 & IL & 98.6 & 98.6 & 95.5 & \textbf{100} & 91.3 & 93.5 \\
    DrivoR~\cite{kirby2026drivor} & CVPR'26 & IL & 98.9 & 98.3 & 96.2 & \textbf{100} & 89.1 & 93.1 \\
    RAP-DINO$^\S$~\cite{feng2026rap} & ICLR'26 & IL & \textbf{99.1} & 98.9 & 96.7 & \textbf{100} & 90.3 & 93.8 \\
    \midrule
    \textbf{\ours (ours)} & -- & IL & 98.9 & 99.0 & 96.3 & \textbf{100} & \textbf{91.5} & \textbf{94.2} \\
    \bottomrule
  \end{tabular}
  \end{adjustbox}
  \caption{\textbf{NAVSIM-v1 \texttt{navtest}.} $^\S$Trained on $10{\times}$ more data.}
  \label{tab:navsim_v1}
\end{table}

\begin{table}[htbp]
  \centering\small
  \setlength{\tabcolsep}{4pt}
  \begin{adjustbox}{max width=\linewidth}
\begin{tabular}{l l c c}
    \toprule
    Method & Backbone & Type & EPDMS$\uparrow$ \\
    \midrule
    LTF~\cite{chitta2023transfuser} & ResNet-34 & IL & 24.4 \\
    DiffusionDrive~\cite{liao2025diffusiondrive} & ResNet-34 & IL & 27.5 \\
    RAP-DINO$^\S$~\cite{feng2026rap} & ViT-H & IL & 39.6 \\
    GTRS-Dense~\cite{li2025gtrs} & V2-99 & IL & 45.0 \\
    ZTRS~\cite{li2025ztrs} & V2-99 & RL & 48.1 \\
    DrivoR$^\dagger$~\cite{kirby2026drivor} & ViT-S & IL & 48.3 \\
    \midrule
    \textbf{\ours (ours)} & ViT-S & IL & \textbf{48.6} \\
    \bottomrule
  \end{tabular}
  \end{adjustbox}
  \caption{\textbf{NAVSIM-v2 \texttt{navhard}.} $^\dagger$Released checkpoint, evaluated by us. $^\S$$10{\times}$ more data.}
  \label{tab:navsim_v2}
\end{table}

\noindent\textbf{NAVSIM.}
On NAVSIM-v1 (\cref{tab:navsim_v1}), \ours reaches 94.2 PDMS, competitive with the state of the art, without the additional training data of RAP.
It gains +1.1 PDMS over DrivoR, mostly from ego progress (+2.4 EP, the highest in the table) and drivable-area compliance (+0.7 DAC).
On NAVSIM-v2 \texttt{navhard} (\cref{tab:navsim_v2}), it reaches 48.6 EPDMS, above DrivoR (48.3) and the RL-trained ZTRS (48.1); with DrivoR's registers as the glance, the same gaze reaches 50.9 (\cref{tab:ablation_planners}).
In the stage-wise evaluation, it gains progress and compliance in the first stage and keeps the progress gain in the second, which starts from perturbed, synthetically rendered states, but loses on collisions and TTC there.

\begin{table}[htbp]
  \centering\small
  \setlength{\tabcolsep}{4pt}
  \begin{adjustbox}{max width=\linewidth}
\begin{tabular}{l l ccccc}
    \toprule
    Method & Venue & DS$\uparrow$ & SR$\uparrow$ & Effi.$\uparrow$ & Comf.$\uparrow$ & Ability$\uparrow$ \\
    \midrule
    UniAD-Base~\cite{hu2023uniad} & CVPR'23 & 45.81 & 16.36 & 129.21 & 43.58 & 15.55 \\
    VAD~\cite{jiang2023vad} & ICCV'23 & 42.35 & 15.00 & 157.94 & 46.01 & 18.07 \\
    TCP-traj$^\ast$~\cite{wu2022tcp} & NeurIPS'22 & 59.90 & 30.00 & 76.54 & 18.08 & 34.22 \\
    ThinkTwice$^\ast$~\cite{jia2023thinktwice} & CVPR'23 & 62.44 & 31.23 & 69.33 & 16.22 & 37.17 \\
    DriveAdapter$^\ast$~\cite{jia2023driveadapter} & ICCV'23 & 64.22 & 33.08 & 70.22 & 16.01 & 42.08 \\
    DriveTransformer~\cite{jia2025drivetransformer} & ICLR'25 & 63.46 & 35.01 & 100.64 & 20.78 & 38.60 \\
    \midrule
    \textbf{\ours (ours)} & -- & \textbf{72.18} & \textbf{45.00} & \textbf{178.35} & 35.14 & \textbf{45.08} \\
    \bottomrule
  \end{tabular}
  \end{adjustbox}
  \caption{\textbf{Bench2Drive closed loop.} Ability: mean multi-ability success rate (\%). $^\ast$Distilled from a privileged expert.}
  \label{tab:b2d}
\end{table}

\noindent\textbf{Bench2Drive.}
In closed loop (\cref{tab:b2d}), \ours reaches 72.18 DS and a 45.0\% success rate, 8.0 DS above the best compared planner, DriveAdapter, which distills features from a privileged expert, and 26.4 DS above the perception-supervised UniAD.
It has the best mean ability score, with the largest margins in merging and overtaking, where the plan has to check the lane it is about to enter, and it is weakest at giving way.
Its comfort is lower than that of UniAD and VAD, which drive more slowly.

\FloatBarrier
\subsection{Ablation studies}
\label{sec:exp:ablation}

\begin{table}[htbp]
  \centering
  \small
  \setlength{\tabcolsep}{4pt}
  \begin{adjustbox}{max width=\linewidth}
\begin{tabular}{l c cc c c ccccc}
    \toprule
    & Gaze & \multicolumn{2}{c}{Gaze in} & Patches & WOD-E2E & \multicolumn{5}{c}{NAVSIM-v1 \texttt{navtest}} \\
    \cmidrule(lr){3-4}\cmidrule(lr){7-11}
    & target & Decoder & Scorer & / hyp. & RFS$\uparrow$ & NC$\uparrow$ & DAC$\uparrow$ & EP$\uparrow$ & TTC$\uparrow$ & PDMS$\uparrow$ \\
    \midrule
    (a) glance only & -- & -- & -- & 0 & 7.87 & 98.4 & 93.0 & 86.2 & 95.1 & 88.21 \\
    (b) decoder gaze & path & \cmark & -- & 2.5k & 7.93 & \textbf{98.9} & 98.0 & 87.1 & 96.1 & 90.41 \\
    (c) scorer gaze & path & -- & \cmark & 2.5k & 8.02 & 98.5 & 98.1 & 90.1 & 95.3 & 92.63 \\
    (d) \textbf{\ours} & path & \cmark & \cmark & 2.5k & \textbf{8.07} & \textbf{98.9} & \textbf{99.0} & \textbf{91.5} & \textbf{96.3} & \textbf{94.20} \\
    \midrule
    (e) random, same count$^\ast$ & random & \cmark & \cmark & 2.5k & 7.71 & 96.1--96.5 & 92.5--92.6 & 80.1--80.4 & 90.3--90.5 & 83.23 \\
    (f) all blocks & all & \cmark & \cmark & 19.0k & 7.98 & 98.7 & 96.7 & 85.7 & 95.4 & 90.15 \\
    \bottomrule
  \end{tabular}\end{adjustbox}
  \caption{\textbf{Where to look, and where to use the evidence} (WOD-E2E val., NAVSIM-v1 \texttt{navtest}). (a) Glance only (token grid, no gaze); (b)--(d) gaze along the path in the decoder (refinement), in the scorer (ranking), or in both; (e) random blocks, as many per hypothesis as the corridor of (d) covers; (f) all 64 blocks. Patches / hyp.: patches read per hypothesis on WOD-E2E. $^\ast$NAVSIM-v1: mean of three runs (range for sub-scores).}
  \label{tab:ablation_where}
\end{table}

The WOD-E2E ablations are trained with the same schedule and seed and evaluated on the validation split.
We report the RFS of the selected proposal at the final checkpoint, as for every run in this paper.

\noindent\textbf{Where to look.}
\cref{tab:ablation_where} holds the architecture fixed and changes only which patches each hypothesis reads.
Row (e) reads as many patches as the full model (d) but at random blocks, with the same parameters and the same compute, and row (f) reads every patch, about eight times more than (d).
The location matters: random blocks are worse than no gaze at all on both benchmarks (7.71 against 7.87 RFS on WOD-E2E, 83.2 against 88.2 PDMS on NAVSIM-v1).
Evidence that does not belong to the hypothesis is not neutral: it distracts.
Reading all 64 blocks is 0.09 RFS and 4.1 PDMS worse than reading the roughly eight blocks along the path.
The path gaze is therefore not a cheaper approximation of dense attention but a better inductive bias: when every hypothesis reads every block, the gaze carries the same shared context as the glance, and the evidence in which the hypotheses differ is diluted.

\noindent\textbf{Refinement and ranking.}
Rows (a)--(d) separate the two uses of the acquired evidence.
Both help, and ranking helps more: the gaze in the scorer alone (c) gains 0.15 RFS and 4.4 PDMS over the glance-only model (a), two and a half times the RFS gain and twice the PDMS gain of the gaze in the trajectory decoder alone (b, 0.06 RFS and 2.2 PDMS), and the two are complementary (d, 0.20 RFS and 6.0 PDMS).
On NAVSIM-v1, the gaze improves every sub-score from (a) to (d), most of all drivable-area compliance (+6.0 DAC) and ego progress (+5.3 EP), and also collisions (+0.5 NC) and time to collision (+1.2 TTC): without it, the planner leaves the road more often and makes less progress.

\begin{table}[htbp]
  \centering
  \small
  \setlength{\tabcolsep}{4pt}
  \begin{tabular}{l c l l}
    \toprule
    Planner & Gaze & PDMS$\uparrow$ & EPDMS$\uparrow$ \\
    \midrule
    DrivoR (registers) & -- & 93.1 & 48.3 \\
    DrivoR (registers) + gaze & \cmark & 93.8 {\scriptsize(+0.7)} & \textbf{50.9} {\scriptsize(+2.6)} \\
    \textbf{\ours} (token grid + gaze) & \cmark & \textbf{94.2} & 48.6 \\
    \midrule
    DiffusionDrive & -- & 88.1 & 27.5 \\
    DiffusionDrive + gaze & \cmark & 90.1 {\scriptsize(+2.0)} & 30.1 {\scriptsize(+2.6)} \\
    \bottomrule
  \end{tabular}
  \caption{\textbf{The gaze on three global scene representations} (NAVSIM-v1 \texttt{navtest}, NAVSIM-v2 \texttt{navhard}).}
  \label{tab:ablation_planners}
\end{table}

\noindent\textbf{Generality across scene representations.}
The gaze needs only a trajectory and the camera calibration, so we add it unchanged to planners with different global representations (\cref{tab:ablation_planners}).
On DrivoR, the registers remain the glance and the path gaze reads the patch features, which gains 0.7 PDMS on NAVSIM-v1 and 2.6 EPDMS on NAVSIM-v2.
On DiffusionDrive, a diffusion planner on a perception-supervised BEV with a ResNet-34 backbone, the same gaze gains 2.0 PDMS and 2.6 EPDMS.
On our token grid it gains 6.0 PDMS (\cref{tab:ablation_where}, (a) to (d)).
The gaze thus improves all three representations, backbones and decoders, and the choice of glance is secondary: with the gaze, the token grid is 0.4 PDMS better than the registers on NAVSIM-v1 and 2.3 EPDMS worse on NAVSIM-v2, and without it the token grid is the weaker summary (88.2 against 93.1 PDMS), so what \ours gains over DrivoR comes from the gaze, not from its scene tokens.

\begin{table}[htbp]
  \centering
  \small
  \setlength{\tabcolsep}{2pt}
  \begin{adjustbox}{max width=\linewidth}
\begin{tabular}{l ll ll ll}
    \toprule
    & Params & Latency & \multicolumn{2}{c}{GFLOPs} & \multicolumn{2}{c}{Memory (MiB)} \\
    \cmidrule(lr){4-5}\cmidrule(lr){6-7}
    & (M) & (ms) & Total & Head & Weights & Peak \\
    \midrule
    w/o gaze & 40.8 & 56.5 & 1854.6 & 2.2 & 188 & 504 \\
    \ours & 53.6 {\scriptsize(+12.8)} & 68.9 {\scriptsize(+12.4)} & 1892.9 {\scriptsize(+38.3)} & 40.5 {\scriptsize(+38.3)} & 237 {\scriptsize(+49)} & 593 {\scriptsize(+89)} \\
    \bottomrule
  \end{tabular}\end{adjustbox}
  \caption{\textbf{Cost of the gaze} (NAVSIM-v1, $672{\times}1148$, four cameras). Inference at batch size 1 on one H200 GPU, mean latency. The backbone takes 1852.4 GFLOPs in both models.}
  \label{tab:overhead}
\end{table}

\noindent\textbf{Cost of the gaze.}
At inference (\cref{tab:overhead}), the gaze adds 12.8\,M parameters, 12.4\,ms of latency (+22\%) and 89\,MiB of peak memory.
The head grows from 2.2 to 40.5 GFLOPs, but the unchanged backbone dominates, so the total compute grows by only 2.1\%.

\subsection{Qualitative analysis}
\label{sec:exp:qualitative}

\begin{figure}[H]
  \centering
  \includegraphics[width=0.92\linewidth]{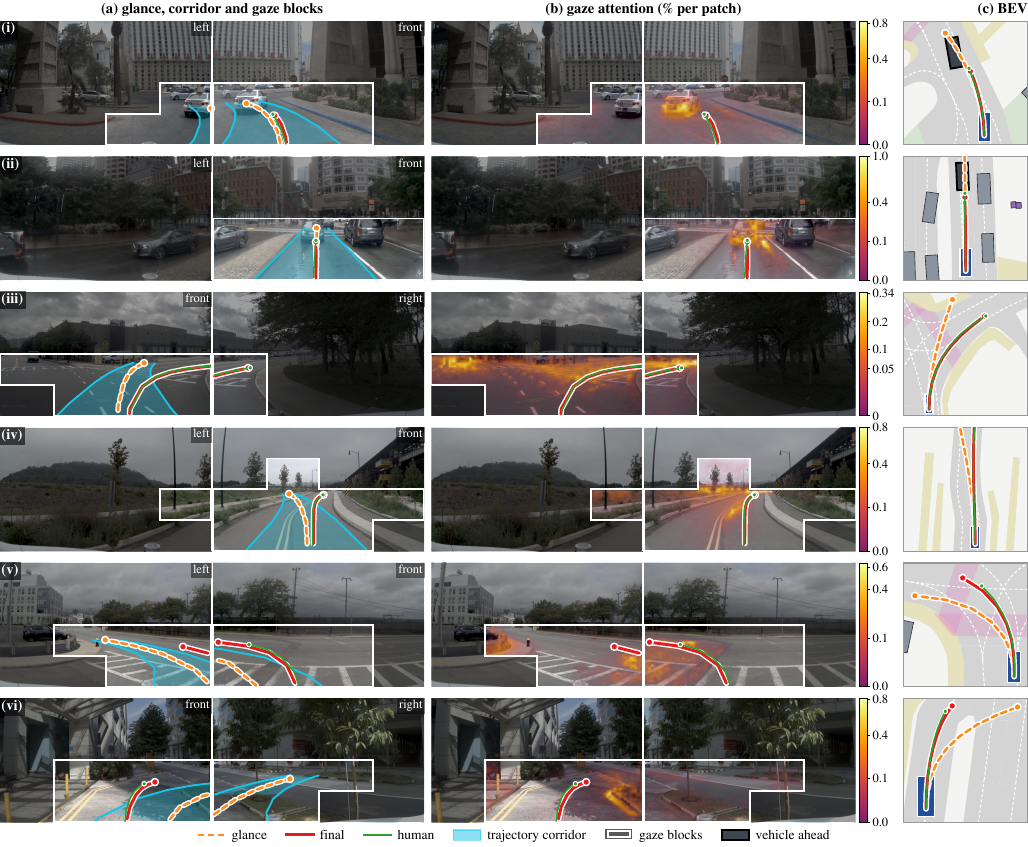}
  \caption{\textbf{The gaze reads the evidence against the glance} (\ours, NAVSIM-v1 \texttt{navtest}).
  (a)~Trajectory of the selected proposal generated in the glance stage, its corridor and the gaze blocks it selects; (b)~gaze attention over these blocks; (c)~bird's-eye view.
  The glance drives into the vehicle ahead (i, ii), misses a right turn (iii), drifts across the centre line (iv) or cuts across the kerb (v, vi); the gaze reads the evidence against it, and the final trajectory follows the human one.}
  \label{fig:qual_navsim}
\end{figure}

\cref{fig:qual_navsim} shows six NAVSIM-v1 scenes in which the trajectory that the glance stage generates for the selected proposal is wrong by 2.3 to 3.4\,m: it drives into a vehicle (i, ii), misses a turn (iii), leaves its lane (iv) or cuts across a kerb (v, vi).
In each case the corridor of this wrong trajectory selects the blocks that contain the evidence against it, the gaze reads that evidence, and the final trajectory returns to within 0.5\,m of the human path.

\noindent\textbf{Limitations.}
The gaze acquires only the evidence that lies along the path, so cues that matter but lie off every path are seen only through the coarse scene tokens.
It also relies on accurate camera calibration, and our closed-loop evaluation is limited to simulation.

\section{Conclusion}
\label{sec:conclusion}

We showed that hypothesis-specific evidence acquisition is a general mechanism for improving perception-free planning.
Instead of letting every candidate trajectory consume the same plan-agnostic summary, each hypothesis gazes at the full-resolution patches along its own projected path, and this evidence refines the hypothesis and, with the larger effect, conditions its ranking.
Nothing about where to look is learned or labeled: the trajectory is the attention mask.
Because the gaze is independent of the global scene representation, it improves planners built on learned registers, on a BEV and on a token grid alike, and \ours achieves the highest WOD-E2E test RFS among the methods in \cref{tab:wod} and is competitive with the state of the art on NAVSIM.
More broadly, our results suggest that better global representations are not the only way forward: a planner with many hypotheses can let each of them decide which evidence to acquire.

\section*{LLM Usage Statement}
The authors used LLM-based assistants (ChatGPT and Claude) to assist with language editing and refinement of author-written text, related work, citations, and code, including the plotting code of the figures. No LLMs or other generative AI tools were used for experimental design or execution, dataset preparation, or the development of research ideas, hypotheses, claims, or conclusions. All LLM-assisted content was reviewed and verified by the authors, who take full responsibility for the submitted work.